\documentclass[conference]{IEEEtran}
\IEEEoverridecommandlockouts
\usepackage{cite}
\usepackage{amsmath,amssymb,amsfonts}
\usepackage{graphicx}
\usepackage{textcomp}
\usepackage{xcolor}

\usepackage{algorithm}
\usepackage{algpseudocode}

\usepackage{multirow}
\usepackage{booktabs}

\def\BibTeX{{\rm B\kern-.05em{\sc i\kern-.025em b}\kern-.08em
    T\kern-.1667em\lower.7ex\hbox{E}\kern-.125emX}}

\begin{document}

\title{Temporal Streaming Batch Principal Component Analysis for Time Series Classification\\
}

\author{\IEEEauthorblockN{1\textsuperscript{st} Enshuo Yan}
\IEEEauthorblockA{\textit{Qingdao Innovation and Development Center} \\
\textit{Harbin Engineering University}\\
Qingdao, China  \\
yanenshuo@hrbeu.edu.cnt}
\and
\IEEEauthorblockN{2\textsuperscript{nd} Huachuan Wang}
\IEEEauthorblockA{\textit{Qingdao Innovation and Development Center} \\
\textit{Harbin Engineering University}\\
Qingdao, China \\
hcwang@hrbeu.edu.cn}
\and
\IEEEauthorblockN{3\textsuperscript{rd} Weihao Xia}
\IEEEauthorblockA{\textit{Qingdao Innovation and Development Center} \\
\textit{Harbin Engineering University}\\
Qingdao, China \\
weihao.xia@hrbeu.edu.cn}
}

\maketitle

\begin{abstract}
In multivariate time series classification, although current sequence analysis models have excellent classification capabilities, they show significant shortcomings when dealing with long sequence multivariate data, such as prolonged training times and decreased accuracy. This paper focuses on optimizing model performance for long-sequence multivariate data by mitigating the impact of extended time series and multiple variables on the model. We propose a principal component analysis (PCA)-based temporal streaming compression and dimensionality reduction algorithm for time series data (temporal streaming batch PCA, TSBPCA), which continuously updates the compact representation of the entire sequence through streaming PCA time estimation with time block updates, enhancing the data representation capability of a range of sequence analysis models. We evaluated this method using various models on five real datasets, and the experimental results show that our method performs well in terms of classification accuracy and time efficiency. Notably, our method demonstrates a trend of increasing effectiveness as sequence length grows; on the two longest sequence datasets, accuracy improved by about 7.2\%, and execution time decreased by 49.5\%.
\end{abstract}

\begin{IEEEkeywords}
time series classification, PCA, temporal compression
\end{IEEEkeywords}

\section{Introduction}
As practical demand continues to grow, the scope of time series data has expanded rapidly in terms of quantity and the increasing complexity of inherent characteristics\cite{b1}, such as its multivariate structure and extended time spans. For example, IoT devices in industries, smart homes, and health monitoring generate large volumes of complex time series data that cover long time spans to analyze trends and make predictions or classifications \cite{b2,b3,b4}. This necessitates the continuous collection, storage, and processing of vast amounts of data with long sequences\cite{b5}. Modern analysis not only focuses on individual variables but also increasingly emphasizes the interactions among multiple variables, thereby placing greater demands on data processing and analytical capabilities\cite{b6}.

\begin{figure}[htbp]
\centering  
\includegraphics[width=0.45\textwidth]{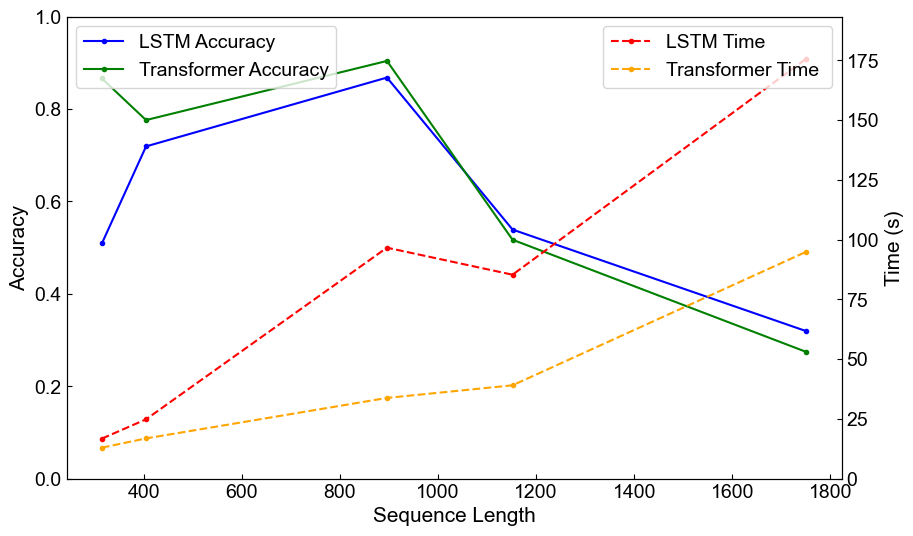}  
\caption{The trends in accuracy and execution time vary with the sequence length.}
\label{fig}
\end{figure}

In the past, many statistical and machine learning methods have been developed for time series analysis, leading to remarkable success. Recurrent neural networks (RNNs)\cite{b7} have played a significant role in sequence modeling. By introducing a recurrent mechanism, RNNs are capable of retaining information from previous steps, making them suitable for handling data with sequential dependencies. However, basic RNNs have the problem of vanishing or exploding gradients when dealing with long sequences. To overcome these issues, gated mechanisms were introduced, such as gated recurrent units (GRU)\cite{b8}  and long short-term memory (LSTM)\cite{b9} .The Transformer processes sequence data concurrently through its encoder-decoder architecture and introduces a self-attention layer, which improves efficiency and performance\cite{b10, b11, b12}, demonstrating superiority over RNNs. However, in intricate temporal patterns, the attention mechanism struggles to find reliable dependencies. Other emerging architectures, such as TimesNet\cite{b13}, decompose temporal variations into multiple periodic components, allowing for a deeper exploration of temporal dependencies. Another category of methods leverages temporal convolutional networks (TCN)\cite{b14, b15}, which utilize convolutional neural networks along the temporal dimension to capture temporal patterns. However, as mentioned earlier, these models generally face limitations when dealing with long sequences, struggling to effectively capture long-term dependencies. This often results in reduced accuracy and increased execution time. Fig. 1 shows the changes in accuracy and execution time of LSTM and Transformer models across varying data lengths, revealing a general decline in performance metrics as the data length increases.

To address the challenges of long time series classification, we propose a method based on streaming PCA that generates compact representations at both the time-step and variable levels, thereby mitigating the effects of long-term dependencies during model training. Traditional PCA methods, such as those based on eigen-decomposition or singular value decomposition (SVD)\cite{b16,b17}, are unsupervised dimensionality reduction techniques for non-time series data that typically compute the covariance matrix of the entire dataset to extract principal components. Based on this, Oja proposed streaming PCA in 1982\cite{b18, b19}, which uses a batch stochastic power method that randomly samples data for PCA computation. This approach avoids the explicit formation of a covariance matrix, offering significant memory advantages. Building on Oja's method, Yang et al.\cite{b20} proposed hist-PCA by incorporating historical information to enhance the algorithm's stability. However, since the hist-PCA algorithm does not account for the temporal dimension, it is still facing challenges in handling time series data effectively. We extend this method to incorporate the temporal dimension, adapting it to the characteristics of time series data. By incrementally updating along the data's temporal flow, our approach retains temporal dependencies while extracting principal directions among multiple variables, resulting in a unique data representation.

We propose a specialized method with three new contributions: (1) We design a novel streaming dimensionality reduction algorithm that utilizes the \textbf{temporal} historical information of data to generate compact representations at both the time-step and variable levels. (2) We systematically analyze the internal parameters of the proposed algorithm, including the time batch size $T$ and the number of retained principal components $K$, to determine their optimal ranges. (3) We demonstrate the applicability of our algorithm across various model architectures, showing that it nearly matches the baseline performance while achieving significant improvements in computational time. Furthermore, we achieved the most outstanding performance improvements on the SCP2 and Ethanol datasets.

\section{OUR METHOD}


\begin{algorithm}
\caption{Temporal Streaming Batch PCA}
\begin{algorithmic}[1]  
\Require A sequence of time series data $\{X_1, \ldots, X_N\}$, Data size $B$, Time batch $T$, starting vector $H \sim N(0, I_{d \times d})$, $H = QR:$ (QR-decomposition)
\Ensure Compact representation of data $Y = \{Y_1, \ldots, Y_t\}$

\While{$Q_1$ not converge}
    \State $W_1 \gets Q_1 + \frac{1}{B} X^T X_1 Q_1$
    \State $W_1 \gets Q_1R_1$
\EndWhile
\State $\lambda_\tau = \|W_1[:\tau]\|_2$ for $\tau = 1, \ldots, k$
\State $\Lambda_1 = \text{diag}(\lambda_1, \ldots, \lambda_\tau)$
\For{$i = 2$ to $N/T$}
    \State $X_i = X[i-1 \times T : i \times T]$
    \For{$j = 1$ to $t$}
        \While{$Q_j$ not converge}
            \State $W_j \gets \frac{j - 1}{j} Q_{j-1} \Lambda_{j-1} Q^T_{j-1} Q_j + \frac{1}{j} \cdot \frac{1}{B} X^T X Q_j$
            \State $W_j \gets Q_j R_j$
        \EndWhile
        \State $\lambda_\tau = \|W_j[:\tau]\|_2$ for $\tau = 1, \ldots, k$
        \State $\Lambda_j = \text{diag}(\lambda_1, \ldots, \lambda_\tau)$
    \EndFor
    \State $Y_i = X_i Q_j$
\EndFor
\State \Return $Y$
\end{algorithmic}
\end{algorithm}

We consider implementing stepwise dimensionality reduction of data through time-dependent estimation based on streaming PCA, focusing on the advantage of streaming PCA in processing data sequentially. By leveraging incremental iteration, each newly arriving data point remains directly linked to the previously received data.Specifically for time series data, our approach involves sequentially feeding data into PCA, rather than randomly sampling independent instances, can fully capture and retain the temporal dependencies of the data, thereby more accurately reflecting the dynamic characteristics of time series and fully accounting for the temporal dependencies in the data.

\setlength{\tabcolsep}{11pt} 
\begin{table*}[t]
\caption{Performance comparison of different methods on various datasets}
\begin{center}
\begin{tabular}{|l|l|c|c|c|c|c|c|c|c|c|c|}
\hline
\textbf{Dataset} & \textbf{Method} & \multicolumn{2}{c|}{\textbf{LSTM}} & \multicolumn{2}{c|}{\textbf{Transformer}} & \multicolumn{2}{c|}{\textbf{Informer}} & \multicolumn{2}{c|}{\textbf{iTransformer}} & \multicolumn{2}{c|}{\textbf{TimesNet}} \\
\cline{3-12}
 & & \textbf{Acc} & \textbf{Time} & \textbf{Acc} & \textbf{Time} & \textbf{Acc} & \textbf{Time} & \textbf{Acc} & \textbf{Time} & \textbf{Acc} & \textbf{Time} \\
\hline
\multirow{2}{*}{UWave} & TSBPCA & \textbf{57.9} & \textbf{9.08} & \textbf{87.2} & 13.12 & 85.6 & \textbf{13.94} & 76.9 & \textbf{12.10} & 82.2 & 20.97 \\
 & w/o PCA & 50.9 & 16.72 & 86.6 & \textbf{12.94} & \textbf{86.2} & 14.40 & 76.9 & 12.48 & \textbf{85.6} & \textbf{19.76} \\
\hline
\multirow{2}{*}{HB} & TSBPCA & \textbf{74.8} & \textbf{16.26} & 76.6 & \textbf{16.20} & \textbf{78.0} & \textbf{17.22} & 75.1 & \textbf{15.35} & 75.1 & 17.84 \\
 & w/o PCA & 71.9 & 24.80 & \textbf{77.6} & 16.80 & 77.6 & 18.13 & \textbf{76.6} & 15.96 & \textbf{76.1} & \textbf{17.49} \\
\hline
\multirow{2}{*}{SR1} & TSBPCA & \textbf{87.5} & \textbf{35.96} & \textbf{90.1} & \textbf{18.27} & 87.7 & \textbf{20.14} & \textbf{85.3} & \textbf{17.82} & 85.3 & 48.60 \\
 & w/o PCA & 86.8 & 96.58 & \textbf{90.4} & 33.73 & \textbf{90.1} & 21.59 & 90.4 & 18.51 & \textbf{90.1} & \textbf{43.69} \\
\hline
\multirow{2}{*}{SR2} & TSBPCA & \textbf{54.8} & \textbf{32.29} & \textbf{57.2} & \textbf{15.50} & \textbf{57.2} & 19.09 & \textbf{59.4} & \textbf{14.68} & \textbf{53.9} & \textbf{42.79} \\
 & w/o PCA & 53.9 & 85.28 & 51.7 & 39.02 & 54.4 & \textbf{18.75} & 54.4 & 16.68 & 51.7 & 55.34 \\
\hline
\multirow{2}{*}{EC} & TSBPCA & \textbf{33.1} & \textbf{57.45} & \textbf{32.3} & \textbf{19.88} & \textbf{30.4} & \textbf{20.20} & \textbf{30.8} & \textbf{19.88} & \textbf{31.2} & \textbf{41.54} \\
 & w/o PCA & 31.9 & 175.68 & 27.4 & 94.92 & 29.7 & 25.56 & 27.0 & 18.81 & 30.0 & 52.86 \\
\hline
\multirow{2}{*}{AVERAGE} & STBPCA & \textbf{61.6} & \textbf{30.21} & \textbf{68.7} & \textbf{16.59} & \textbf{67.8} & \textbf{18.12} & \textbf{65.5} & \textbf{15.97} & \textbf{65.5} & \textbf{35.75} \\
 & w/o PCA & 59.1 & 79.81 & 66.7 & 39.48 & 67.6 & 19.69 & 65.1 & 16.49 & 66.7 & 37.83 \\
\hline
\end{tabular}
\label{tab:performance_comparison}
\end{center}
\end{table*}

The algorithm performs a forward double iteration along the sequential direction of the entire dataset. The proposed method is outlined in Algorithm 1, given a time batch size $T$ and the number of principal components $K$. The time estimation is expressed as $X_i Q_i$, where \(X_i\) is the original data and \(Q_i\) is the corresponding temporal projection matrix.

At time step 1, we receive the first time batch \(X_1\) of the data. For the first time point within batch \(T\), since no past temporal information is available, we begin by using a randomly initialized \(d \times k\) matrix \(Q_1\) to find the matrix of \(k\) feature vectors for the first time point in this batch, denoted as \(W_1\): 
\[
W_1 \leftarrow Q_1 + \frac{1}{B} X_1^T X_1 Q_1 \tag{1}
\]
Where \( \frac{1}{B} X_1^T X_1 \) represents the sample covariance matrix composed of the \(B\) data points that enter the PCA at the first time point, as shown in line 2 of Algorithm 1. In this context, data from a specific time point across the entire dataset is used. A larger volume of data during each iteration can help reduce potential high variance by averaging out noise\cite{b20}.

At time step 2, we begin processing the data \(X_j\) for the next time point within the current batch. Unlike in time step 1, we now incorporate past temporal information. By combining the most recent representation \(W_1\) updated in the previous step, we can use the estimator \( \frac{1}{2} \left( \frac{1}{B} X_2^T X_2 + W_1 W_1^T \right) \), which takes into account historical information. Extending to time step \(j\) (\(1 < j < t\)), the latest estimate of the covariance matrix can be obtained by leveraging the history from the previous \(j - 1\) steps. Intuitively, there is no reason to favor any particular time point, so we assign equal weights to the data within each time batch, giving weights of \((j - 1)/j\) and \(1/j\). The updated matrix \(W_j\) is then given by:

\[
W_j \leftarrow \frac{j-1}{j} Q_{j-1} \Lambda_{j-1} Q_{j-1}^T + \frac{1}{j} \frac{1}{B} X_j^T X_j Q_j \tag{2}
\]
Line 11 of Algorithm 1 describes this process. \(Q_j\) is the matrix of \(K\) eigenvectors, and \(\Lambda_j\) is the \(K \times K\) diagonal matrix where the diagonal elements correspond to the eigenvalues at time \(j - 1\).

At time \(t\) within the current time batch, after updating the latest representation, the batch output value \(Q_i\) is used as the time projection matrix for the batch data to complete the compact representation for this time point. \(Q_i\) will then replace the initial random values as the prior historical information for the next batch, participating in a new round of inner iterations. This process continues until all time points in the data are reached, resulting in the desired compact representation of the data.

\section{EXPERIMENTS AND RESULTS}
Our experiments aim to evaluate the effectiveness of the TSBPCA method by comparing the differences in execution time and accuracy achieved by TSBPCA against the baseline methods.

\subsection{Experiment Setup}\label{AA}
We selected five real-world datasets with varying sequence lengths: the UWaveGestureLibrary dataset (UWave)\cite{b21}, the Heartbeat dataset (HB)\cite{b22}, the SelfRegulationSCP1 dataset (SCP1)\cite{b23}, the SelfRegulationSCP2 dataset (SCP2)\cite{b23}, and the EthanolConcentration dataset (EC)\cite{b24}. These datasets contain variables ranging from 3 to 61 and sequence lengths from 300 to 1800, allowing for a comprehensive evaluation of various sequence lengths. We evaluated our method across five different models: LSTM, Transformer, Informer, iTransformer, and TimesNet. All experiments were conducted on a platform equipped with a single NVIDIA GeForce RTX 3090 GPU. For each experiment, all models were configured with dataset-specific pre-selected default hyperparameters to achieve optimal performance. The batch size for all experiments was set to 16.

\subsection{Experiments on Different Parameters}
We analyzed the parameters \(T\) and \(K\) of the TSBPCA algorithm and conducted tests using the Heartbeat dataset. By setting different values for \(T\) and \(K\), we observed changes in accuracy, revealing the following findings: (1)The value of \(K\) is closely related to the amount of data. When \(K=1\), the model accuracy is 74.8\%. However, when \(K\) increases to 2, the accuracy drops to 69.8\%. More importantly, when \(K\) is greater than 3, the training becomes unstable, and the loss turns into invalid values. This is due to the positive correlation between the PCA method's approximation accuracy and the number of samples. When the amount of data is small, the error in the PCA method increases, resulting in training instability. (2)As \(S\) increases, the model's accuracy tends to decrease. Fig. 2 illustrates the impact of \(S\) on the results, suggesting that \(S\) should be chosen by balancing model execution time and accuracy. For the Heartbeat dataset, the optimal range for \(S\) is between 2 and 5, with the best value being 3. The specific \(S\) value for different datasets needs to balance between model running speed and accuracy to select the best performance.

\subsection{Performance}
The experimental results are summarized in Table~\ref{tab:performance_comparison}, which compares the results obtained using the TSBPCA algorithm with those obtained without using enhancement methods. For each entry in Table~\ref{tab:performance_comparison}, the internal parameters \( T \) and \( K \) of the TSBPCA algorithm are the values that yielded optimal performance after multiple experiments. The experimental results can be summarized as follows:

\begin{figure}[htbp]
\centering  
\includegraphics[width=0.45\textwidth]{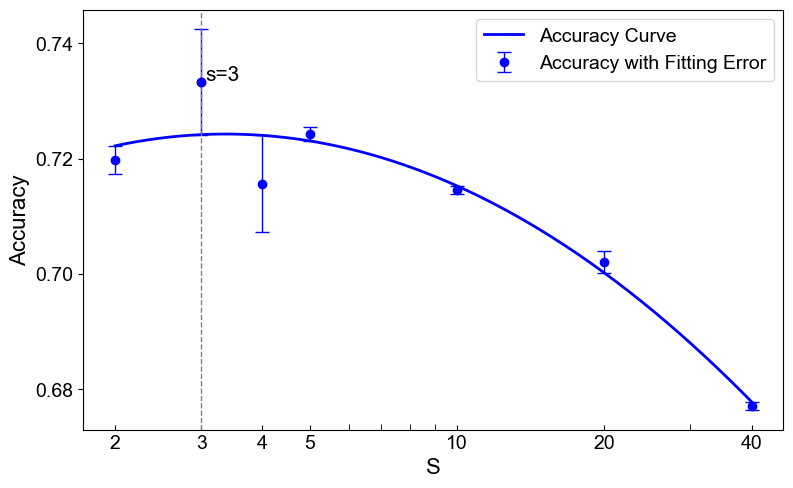}  
\caption{Accuracy trends with respect to parameter S on a log scale.}
\label{fig}
\end{figure}

\begin{itemize}
\item For the RNNs models, the TSBPCA + LSTM model demonstrates a significant performance improvement across all five datasets. The accuracy improves by an average of 4.23\%, and the time efficiency improves by an average of 62.15\%. This enhancement is attributed to the RNN's inherent adaptability to sequential data due to its cyclic structure, allowing it to effectively adapt to our method.
\item For other TSBPCA-enhanced models, the improvement is somewhat less pronounced compared to the RNN models. Results vary across different datasets and do not consistently show better performance. However, on average, these models still generally outperform conventional methods in terms of both accuracy and time efficiency.
\item From a data perspective, there is a significant variation in the improvement effects across different datasets. Shorter sequence lengths show less improvement compared to longer sequences. The method still achieves notable improvements on the SR2 and EC datasets, with accuracy improving by up to 18\% on average and time expenditure decreasing by up to 80\% on average. This indicates that the method is well-suited for long sequence data.

\item In addition, the size of the dataset also affects the experimental results. Currently, large-scale datasets in multivariate time series are relatively scarce, which limits further exploration of the relationship between dataset size and the effectiveness of the method. However, it is evident that the acceleration effect of our method becomes more pronounced with larger datasets. It can be anticipated that as the dataset size increases, our method will show more significant improvements, not only in terms of enhanced accuracy in time estimation but also in reflecting the Inverse relationship between execution speed and dataset size.
\end{itemize}

\section{CONCLUSIONS}
In summary, we introduce the TSBPCA method that generates low-dimensional approximations of data through time estimation in streaming PCA, significantly optimizing the performance of multivariate time series classification models when handling long sequences. The experimental results demonstrate that the method improves classification accuracy and efficiency across multiple real-world datasets and models, particularly showing stronger adaptability to long sequence data. Our approach effectively preserves key features in sequence data while reducing unnecessary computational overhead, making it suitable for more complex real-world scenarios.Moving forward, we plan to explore more complex algorithm implementations, such as improving the dual-iteration internal data reconstruction mechanism.



\vspace{12pt}

\end{document}